# Performance Evaluation of DCA and SRC on a Single Bot Detection

Yousof Al-Hammadi, Uwe Aickelin and Julie Greensmith

Intelligent Modelling & Analysis (IMA) Group
School of Computer Science, The University of Nottingham,
Wollaton Road, NG8 1BB, Nottingham, UK
*{yxa,uxa,jqg}@cs.nott.ac.uk*

***Abstract:*** Malicious users try to compromise systems using new techniques. One of the recent techniques used by the attacker is to perform complex distributed attacks such as denial of service and to obtain sensitive data such as password information. These compromised machines are said to be infected with malicious software termed a "bot". In this paper, we investigate the correlation of behavioural attributes such as keylogging and packet flooding behaviour to detect the existence of a single bot on a compromised machine by applying (1) Spearman's rank correlation (SRC) algorithm and (2) the Dendritic Cell Algorithm (DCA). We also compare the output results generated from these two methods to the detection of a single bot. The results show that the DCA has a better performance in detecting malicious activities.

## 1. Introduction

Computer systems and networks come under frequent attack from a diverse set of malicious programs and activity such as viruses and worms [10]. The detection of such threat is improving in the area of network and computer security. Recently, a new threat has emerged in the form of the *botnet*. Botnets, which are groups of distributed bots, are controlled remotely by a central commander, termed the "botmaster". A single bot, a term derived from robot, is a malicious piece of software which, when installed on a compromised host, transforms host into a zombie machine. This zombie machine is remotely controlled by the attacker.

Bots use different types of networking protocols for the communication component of their Command and Control (C&C) structure such as Internet Relay Chat (IRC), HTTP and more recently Peer-to-Peer (P2P). In this research we are primarily interested in the detection of bots which use IRC protocol as they appear to be highly prevalent within the botnet community. IRC [17] is a chat based protocol consisting of various "channels" to which a user of the IRC network can connect. The attacker programs his bots to connect to the IRC server and joins the specified channel waiting for his commands. Once the attacker joins the same channel, he starts to issue various commands and all available bots on that channel respond to these commands through C&C structure. In early implementations, bots were used to perform distributed denial of services attacks (DDoS) using a flood of TCP SYN, UDP or ICMP "ping" packets in an attempt to overload the capacity of computing resources.

Recent bots are developed complete with advance features such as keylogging for closely monitoring user behaviour including the interception of sensitive data such as passwords, monitoring mouse clicks and the taking of screenshots of secure websites. Many Anti-Virus packages cannot detect a stealthy keylogging activity on the system. The user has no way to determine if his machine is running a keylogger, therefore, he could easily become a victim of the identity theft.

Many existing botnet/bot techniques use different types of signatures-based detection by analysing network traffic in order to detect botnets as in [6][9][20]. These detection techniques can be evaded by either changing the bot's signatures or encrypting the bot's traffic when communicating with the attacker. In addition, a bot can connect to non-standard ports to make the detection more difficult. Rather than detecting botnet by monitoring and analysing network traffic looking for bots' signatures, our work focuses on the detection of a single bot formulated as a host-based intrusion detection problem, and avoid the technical problems of administrating a highly infective network within an academic environment. To perform this research, we rely on principles of "extrusion detection" where we do not attempt to prevent the bot from gaining access to the system, but we detect it as it attempts to operate and subvert the infected host. This procedure involves monitoring different bot's behaviours within specified time window such as potential keylogging activity and fast reaction to the received network information.

In order to detect the bot on the infected machine, correlating bots' behavioural attributes is needed. The concept of correlation attributes within specified time-window increases the level of malicious behaviour activities as depending on one process attribute may generate large number of false alarms. This is also lead to the challenge of choosing the right correlation algorithm which enhances the detection of malicious program.

In previous work [2][3], we introduce two different algorithms to correlate the behaviour of the bot running on the infected system. In this work, we compare and evaluate the performance of the two correlation algorithms on bot detection, including Spearman's rank correlation (SRC) and the DCA. The SRC algorithm examines the correlation of different processes behaviours by monitoring specified function calls executed by running processes on a single machine. DCA has been applied to many problems particularly in the area of intrusion detection in computer security. The DCA is a more intelligent way of fusing and correlating information from disparate sources. The immune inspired DCA implemented by Greensmith et at. [12] is based on an abstract model of the behaviour of dendritic cells (DCs) [22]. These cells are the natural intrusion detection agents of the human body, which activate the immune system in response to the detection of damage to host tissues. As an algorithm, the DCA performs multi-sensor data fusion on a set of input "*signals*", and this information is correlated with potentially anomalous "suspect entities" which we term "*antigen*". This results in information which will state not only if an anomaly is detected, but in addition the culprit responsible for the anomaly. Given the success of this algorithm at detecting scanning activity in computer networks as in [13][14], we will examine the DCA as a solution to correlate different behaviours of a single bot running on a machine.

The aim of this paper is to investigate the effect of correlating bot's behavioural attributes by applying two specified correlation algorithms to the detection of a single bot. For these experiments the basis of classification is facilitated through the correlation of different activities such as keystrokes interception, how fast the program executes certain communication function calls and how fast is the program react when receiving information. Our results show that correlating behaviours exhibited by a single bot can enhance the detection of malicious processes on the system to determine the presence of a bot infection and to identify the processes involved in the bot's actions.

This paper is structured as follows: Section two discusses existing bot detection techniques. Section three describes detection methods that are used to detect a single bot on the system. We present our methodology of bot detection and explain the conducted experiments in section four. Our results and analysis are presented in section five and we summarize and conclude in section six.

## 2. Related Work

Existing research conducted in bot detection concentrates on detecting botnets rather than an individual bot as noted by [1][7][8]. The majority of these techniques use signature-based approaches for botnet detection by analysing network traffic looking for well know signatures. Although this approach is a useful mechanism for bot detection, it is limited if the network packet data is encrypted.

Previous work presented by Barford [4] represents a good introduction to understanding and analysing the behaviours of bots. Freiling et al. [8] collect bot binaries by using a non-productive resource (honeypot), to analyse bot traffic and infiltrate botnet by emulating bot activities.

Cooke et al. [7] performs bot detection through payload analysis using pattern matching of known bot commands and in addition examines a system for evidence of non-human characteristics. While they suggested that correlating data from different sources would be beneficial for the detection of a single bot, they did not provide information regarding how this correlation should be performed. Goebel and Holz [9] monitor and classify IRC traffic based on suspicious IRC nicknames, IRC servers and non-standard server ports using regular expressions.

Anomaly detection plays an important rule on detecting the presence of a bot [5], where deviations from a defined "normal" are classed as an anomaly. An approach for detecting bots using behavioural analysis is presented by Racine [19] which classifies inactive clients and their subsequent assignment to a network connection.

Gu el at. introduce the BotHunter [15], which examines the behaviour history of each distinct host to find correlated evidence of malware infection and the BotSniffer [16], which correlates common bot activities such as coordinated communication, propagation and attack in network traffic.

In summary the majority of techniques for the detection of a single bot uses signature-based detection by analysing network packets. These techniques are limited in case if packet streams are encrypted. Current behavioural-based approaches are also limited, generating high rates of false positives, which have the potential to slow down or denial of service a system. We believe that correlating relevant behavioural attributes with programs potentially involved with a bot infection can enhance the detection mechanism.

## 3. Bot Detection Methods

Existing research techniques detect the presence of bots via network monitoring and analysis. Rather than attempting to detect bots via network analysis, our work focuses on detecting an individual bot running on a machine by monitoring and correlating different activities on the system. In this section, we will describe two algorithms which apply correlation techniques to detect abnormal behaviour in our system.

### 3.1 Spearman's Rank Correlation - SRC

The Spearman's rank correlation (SRC) algorithm to detect the bot is described in Algorithm 1.

```
S₁: keystrokes interception
S₂: how fast the bot responds to attacker
commands
S₃: how fast the bot repeats the same
communication function calls

if (KeyboardState function(s) is executed /*
i.e. keylogging activity*/ )
{
if (SRC(S₁,S₃)>Threshold && SRC(S₂,S₃)>Threshold)
{
    Strong Detection
}
 elseif (SRC(S₁,S₃)<Threshold && SRC(S₂,S₃)<
 Threshold)
 {
    Weak Detection
 }
 elseif ((SRC(S₁,S₃)<Threshold && SRC(S₂,S₃)>
 Threshold) || (SRC(S₁,S₃)>Threshold &&
 SRC(S₂,S₃)< Threshold))
 {
    Medium Detection
 }
}
else
   No detection and normal activity is
   considered
end
```

*Algorithm 1.* SRC Algorithm for detecting Bot.

SRC is a statistical measure of correlation which uses threshold function to describe the relationship between two variables. In order to detect a bot in a system, different bot behaviours are correlated to generate a high correlation value represented by SRC value. Such behaviours include intercepting user keystrokes, how fast the bot responds to the attacker commands and how fast it executes same function calls. In our case, if SRC value exceeds a certain threshold level, a high correlation between the two different behaviours is generated. According to SRC algorithm, the threshold level of 0.5 or higher represents a strong correlation between two events.

The aim of SRC experiments is to verify the notion that correlating different behaviours of a single process indicates abnormal activity. In addition, we apply the monitoring and correlation scheme to a normal application to verify that the normal application behaves differently from the malicious process which results in having different correlation value. The obtained results are compared with DCA results.

### 3.2    The Dendritic Cell Algorithm - DCA

#### 3.2.1    Algorithm Overview

Artificial Immune Systems (AIS) are algorithms inspired by the behaviour of the human immune system. The biological immune system tries to protect the body from the attack against any invading pathogen, viruses and bacterias. AIS have been applied to problems in computer security since their initial development in the mid-1990's.

A recent addition to the AIS family is the Dedritic Cell Algorithm (DCA) implemented by Greensmith et al. [2]. DCA is inspired by the function of the Dendritic Cells (DCs) of the innate immune system and uses principles of a key novel theory in immunology termed the danger theory described by Matzinger [18]. The danger theory suggests that the DCs are the first line defense against invaders and the response is generated by the immune system upon the receipt of molecular information which indicates the presence of stress or damage in the body. The interested reader can refer to [11] for a detailed description of the DCA. In this section we provide an overview of the operation of the algorithm.

When viewed from a computational prospective, DCs are anomaly detector agents, which are responsible for data fusion and generating appropriate actions in response to the attack in the human body. In nature DCs exist in one of three states: *immature*, *semi-mature* and *mature*. The initial maturation state of a DC is immature for sensing and processing three categories of input signals (see Table 1) and in response produces three output signals. The three input signals can influence the behaviour of DCs sensitivity.

The first two input signals are $S_1$ and $S_2$. $S_1$ signal is derived from the detection of pathogens while $S_2$ signal is generated from the unexpected cell death of damage to the tissue cells.

The third input signal is $S_3$ which is molecules released as result of normal cell death. During immature lifespan collecting signals, if the DC has collected majority of $S_3$, it will change state to a semi-mature state and suppress the activation of the immune system. Conversely, cells exposed to $S_1$ and $S_2$ signals transforms into a mature state and can instruct the immune system to activate.

| Signal Name | Symbol | Definitions |
|---|---|---|
| Pathogen Associated Molecular Patterns | $S_1$=PS | A strong evidence of abnormal/bad behaviour. An increase in this signal is associated with a high confidence of abnormality. |
| Danger Signal | $S_2$=DS | A measure of an attribute which increases in value to indicate deviation from usual behaviour. Low values of this signal may not be anomalous, giving a high value confidence of indicating abnormality. $S_2$ has less effect on the output signal than $S_1$ signal. |
| Safe Signal | $S_3$=SS | A measure which increases value in conjunction observed normal behaviour. This is a confident with indicator of normal, predictable or steady-state system behaviour. This signal is used to counteract the effects of $S_1$ and $S_2$ signals and thus has negative impact on the output signals. |

*Table 1.* Signals Definition

While in immature state, DCs capture the suspect entities (termed "antigen") and combine them with evidence of damage in the form of signals to provide information about how "dangerous" a particular protein is to the host body. Antigen collected by the semi-mature DCs are presented in a "safe" context while antigen presented by mature DCs are presented in a "dangerous" context.

In terms of the algorithm, the DCA is a population based algorithm which performs anomaly detection based on the indication of abnormality of the system by aggregating and performing asynchronous correlation of signals with the suspects antigen. Signal processing occurs within DCs of the immature state. Each DC in the immature state performs three functions as follows:

- To sample antigen by collecting antigen from an external source and transfers the antigen to its own antigen storage facility.
- To update input signals in which the DC collects values of all input signals present in the signal storage area.
- To calculate temporary output signal values from the received input signals, with the output values then added to form the cell's cumulative output signals.

The transformation from input to output signal per cell is performed using a simple weighted sum (Equation 1) described in detail in [14] with the corresponding weights given in Table 2 (WS$_3$). These weights determine the value of the output and derived from preliminary observation that defines the danger level of the input signals.

$$O_i = \sum_{i=1}^{3} (W_{ijk} * S_i) \cdots \forall jk \quad (1)$$

Where:
- $W$ is the signal weight of the category $i$
- $i$ is the input signal category ($S_1$=PS, $S_2$=DS and $S_3$=SS)
- $k$ is the weight set index $WS_k$ as shown in Table 2 ($k$ =1 to

5)
- $O_j$ is the output concentrations of one of the following signal:
  - j=1 costimulatory signal (csm)
  - j=2 a semi-mature DC output signal (semi)
  - j=3 mature DC output signal (mat)

| | Signal | WS₁ | WS₂ | WS₃ | WS₄ | WS₅ |
|---|---|---|---|---|---|---|
| | $S_1$ | 2 | 4 | 4 | 2 | 8 |
| O1(csm) | $S_2$ | 1 | 2 | 2 | 1 | 4 |
| | $S_3$ | 2 | 6 | 3 | 1.5 | 0.6 |
| | $S_1$ | 0 | 0 | 0 | 0 | 0 |
| O2(semi) | $S_2$ | 0 | 0 | 0 | 0 | 0 |
| | $S_3$ | 1 | 1 | 1 | 1 | 1 |
| | $S_1$ | 2 | 8 | 8 | 8 | 16 |
| O3(mat) | $S_2$ | 1 | 4 | 4 | 4 | 8 |
| | $S_3$ | -3 | -12 | -6 | -6 | -1.2 |

*Table 2.* Weight Sensitivity Analysis.

In the algorithm, the signal values are assigned real valued numbers and the antigen are assigned as categorical values of the object to be classified. The algorithm has three different stages, the initialization stage, the data processing and the analysis stage. In the initialization stage, the algorithm generates DCs population where each cell is assigned a random "migration" threshold. The input data forms the sorted antigen and signals ($S_1$, $S_2$ and $S_3$) with respect to the
time and passed to the processing stage. Each DC performs an internal correlation between signals and antigen with respect to a specified time window determined by the migration threshold, signals and antigen. To cease data collection, a DC must have experienced signals, and in response to this express output signals. As the level of input signal experienced increases, the probability of the DC exceeding its lifespan also increases. The level of signal input is mapped as a cumulative $O_1$ value. Once $O_1$ exceeds a migration threshold value, the cell ceases signal and antigen collection and is removed from the population and enters the maturation stage. Upon removal from the population the cell is replaced by a new cell, to keep the population level stable.
A high concentration of $S_1$ and $S_2$ increases the probability of
immature cells to become mature cells while a more concentration of $S_3$ imposes the immature cells to become semi-mature cells. Therefore, if $O_2 > O_3$, the DC is termed "semi-mature" cell. Antigen presented by semi-mature cell is assigned a context value of *zero*. In contrast, $O_2 < O_3$ leads to
a "mature" cell and antigen presented by mature cell is assigned a context value of *one*. The detection of anomaly is based on having more mature cells than semi-mature cells in which the antigen in a mature context is detected. The pseudo code for the functioning of a single cell is presented in Algorithm 2.
The final stage involves calculating an anomaly coefficient per antigen type - termed the *mature context antigen value, MCAV* once all antigen and signals are processed by the cell population, an analysis stage is performed. The derivation of the MCAV per antigen type in the range of zero to one is shown in Equation 2.

```
input: Sorted antigen and signals
(S₁=PS,S₂=DS,S₃=SS)
output: Antigen and their context (0/1)

Initilize DC;

foreach cell in DC population
{
   while CSM output signal (O₁) < migration
   threshold
   {
      get antigen;
      store antigen;
      get signals;
      calculate interim output signals;
      update cumulative output signals;
   }
   cell location update to lymph node;

   if semi-mature output (O₂) > mature output(O₃)
      cell context is assigned as 0 ;
   else
      cell context is assigned as 1 ;

   kill cell;
   replace cell in population;
}
```

*Algorithm 1.* DCA Algorithm for detecting Bot.

The closer this value is to one, the more likely the antigen type is to be anomalous. A threshold is applied to distinguish between anomalous and normal type of antigen.

$$MCAV_x = \frac{Z_x}{Y_x} \quad (2)$$

Where $MCAV_x$ is the MCAV coefficient for antigen type $x$, $Z_x$ is the number of mature context antigen presentations for antigen type $x$ and $Y_x$ is the total number of antigen presented for antigen type $x$.
Previously in [11], it has been shown that the MCAV for processes with low numbers of antigen per antigen type generates false positives alarms. In order to reduce these false alarms, we introduced an anomaly value which is an improvement on the MCAV, by incorporating the number of antigen used to calculate the MCAV. This improvement is termed the *MCAV Antigen Coefficient, MAC*. The MAC value is calculated from Equation 3 and also ranges between zero and one. As with the MCAV, the closer the MAC value to one, the more anomalous the process.

$$MAC_x = \frac{MCAV_x * Antigen_x}{\sum_{i=1}^{n} Antigen_i} \quad (3)$$

Where $MCAV_x$ is the MCAV value for process $x$ and $Antigen_x$ is the number of antigen processed by process $x$.

## 4. Methodology

### 4.1 Overview

For the purpose of experimentation two different types of bots are used, namely spybot [4] and sdbot [21]. These are suitable candidate bots as they use a range of malicious functionalities such as keylogging, SYN attack and UDP attack which are frequently used features by bots. An IRC client (IceChat) is used for normal conversation and to send files to a remote host which represents normal traffic. To provide suitable data a "hooking" program is implemented to capture the required behavioural attributes by intercepting specified function calls. The collected data are processed by both the SRC algorithm and the DCA to measure the detection performance.

### 4.2 Bot Scenarios

Three different scenarios are constructed including inactive (E1), attack (E2.1-2.3) and normal (E3) scenarios. The attack scenario consists of three sessions: a keylogging attack session, a flooding session and a combination session comprising both keylogging and packet flooding.

- Inactive bot (E1): The bot on the infected host connects to an IRC server and joins a specified channel to await commands from its controller, though no attacking actions are performed by this idle bot. Other normal applications such as an IRC client, Wordpad, Notepad and terminal emulator (CMD) processes are also running on this host.

- Keylogging Attack (E2.1): The bot is capable of intercepting keystrokes using various methods. Two methods of keylogging are used including the "GetKeyboardState" (E2.1.a) and "GetAsyncKeyState" (E2.1.b) function calls. However, detection cannot be performed by monitoring these function calls alone, as some of legitimate programs often rely on such function calls.

- Flooding Attack (E2.2): This involves performing packet flooding using the spybot for a SYN flood attack (E2.2.a) and the sdbot for a UDP attack (E2.2.b) which emulate the behaviour of a machine partaking in a distributed denial of service attack.

- Combined Attack (E2.3): In this session, both keylogging and flooding attack (SYN flood [E2.3.a] and UDP flood [E2.3.b]) are invoked by the bot. Note that the two activities can occur simultaneously in this scenario.

- Normal Scenario (E3): This involves having normal conversation between the two parties. It also includes transferring a file of 10 KB from one host to another through IRC client. Other applications such as Wordpad, Notepad, cmd and the hook program are running on the victim host. Note that no bots are used in this scenario.

### 4.3 Signals

Three signal categories are used to define the state of the system namely $S_1$, $S_2$ and $S_3$ as described previously in Table 1, with one data source mapped per signal category. The mapping of raw signals to signals for the algorithm is determined via expert knowledge. These signals are collected using a function call interception program. Raw data from the monitored host are transformed into log files, following a signal normalisation process. The resultant normalised signals are in the range of $0 - 100$.

In terms of the signal category semantics, $S_1$ (PS) is a strong evidence for bad behaviour on a system. Because we focus on detecting bots performing keystrokes interception in combination with other malicious activities, we have classified this activity as our $S_1$. This signal is derived from the rate of change of invocation of selected API function calls used for keylogging activity. Such function calls include GetAsyncKeyState, GetKeyboardState, GetKeyNameText and keybd_event when invoked by the running processes. To use this data stream as signal input, the rate values are normalised. For this process $n_{ps}$, (ps is referred to the PS signal), is defined as the maximum number of function calls generated by pressing a key within one second.

$S_2$ (DS) is derived from the time difference between receiving and sending data through the network for each process by intercepting the send() and recv() function calls. Because bots respond directly to botmaster commands, a small time difference between sending and receiving data is observed. In contrast, normal chat between users will have a higher response time. As with $S_1$ signal, the normalisation of $S_2$ involves calculating a maximum value. For this purpose $n_{ds}$, (ds is referred to the DS signal), is the maximum time difference between sending a request and receiving a feedback. If the time difference exceeds $n_{ds}$, the response time is normal. Otherwise, the response time falls within the abnormality range.

We set up a critical range (0 to $n_{ds}$) that represents an abnormal response time. The zero value is mapped to 100 *max-danger* time and $n_{ds}$ is mapped to zero *min-danger* time. If the response time falls within the critical value, it means that the response is fast and considered to be dangerous.

Finally, $S_3$ (SS) is derived from the time difference between two outgoing consecutive communication functions such as [(send,send),(sendto,sendto),(socket,socket),(connect,connect]. This observation is based on bot sending information to the botmaster or issues SYN or UDP attacks which generates many function calls within a short time period. Therefore we set $n_{ss1}$ and $n_{ss2}$ (ss is referred to SS signal) as a range of a time difference between calling two consecutive communication functions. If the time difference is less than $n_{ss1}$, the time is classified within a *min-safe* time. If the time difference falls between $n_{ss1}$ and $n_{ss2}$, the time is classified as *uncertain* time. If the time difference is more than $n_{ss2}$, the time is classified as *max-safe* time. By recording the time that a bot responds to the command in most of the experiments that we have conducted, we notice that the mean value for bot to respond to the command is around 3.226 seconds. Therefore, we set up a critical range for $S_3$ signal. We divide our critical range into three sub-ranges. The first range is from zero to $n_{ss1}$ where $n_{ss1}=5$ to allow enough time for a bot to respond to the attack's command. Any value that falls within this range is considered as a *min-safe* time. The second range is where there is uncertainty of response. The uncertainty range is between $n_{ss1}$ and $n_{ss2}$ =20. The third range is that the time difference is above $n_{ss2}$ and is considered as a *max-safe* time. In this range, we are sure that the time difference between two consecutive function calls is generated as a normal response.

In case of $S_2$ and $S_3$ signals, this decision is based on that the attacker design the bot to responds to the his/her command without adding a short random delay when responding to the commands or when flooding other hosts or network.

### 4.4 Antigen

For the purpose of bot detection, antigen are derived from API function calls, which are similar to system calls. The resultant data is a stream of potential antigen suspects, which are correlated with signals through the processing mechanisms of the DC population. One constraint on antigen is that more than one of any antigen type must be used to be able to perform the anomaly analysis with the DCA. This will allow for the detection of which type of function call is responsible for the changes in the observed input signals.

The collected signals are a reflection of the status of the monitored system. Therefore, antigen are potential culprits responsible for any observed changes in the status of the system. The correlation of antigen signals is required to define which processes are active when the signal values are modified. Any process executed a specified function calls, the process id which causes the calls is stored as an antigen in the antigen log file. The more active the process, the more antigen it generates. Each intercepted function call is stored and is assigned the value of the process ID to which the function call belongs and the time at which it is invoked.

For the SRC algorithm experiments, only the signals ($S_1$, $S_2$, $S_3$) log file is used to detect the malicious activities. In case of DCA, signal and antigen logs are combined and sorted based on time. The combined file forms a dataset which is passed to the DCA through a data processing client. The combined log files are parsed and the logged information is sent to the DCA for processing and analysis.

### 4.5 Data Collection

A bot is already installed on the victim host, through an accidental "trojan horse" style infection mechanism and runs as a process whenever the user reboots the system.

An interception program is implemented and run on the victim machine to collect the required data. Two types of log files are produced, SigLog and AntigLog. The SigLog presents values $S_1$, $S_2$ and $S_3$ in the following format with an example below it:

```
        <time> < type > <S1> <S2> <S3>
   e.g. <0001> <signal> <11> <32> <89>
```

The AntigLog presents the intercepted API function calls with respect to its process ID (PID) in the following format with an example below it:

```
        <time> < type > <PID> <Function call name>
   e.g. <0002> <antigen> <722> <GetAsyncKeyStat() >
```

After finishing the data collection, the SigLog is passed to SRC algorithm for analysis. In case of DCA, the SigLog and AntigLog are merged together and sorted with respect to the time and the combined file is passed to the DCA for the analysis.

Three specific types of function calls are intercepted. These function calls are as follows:

- *Communication* functions: socket, connect, send, sendto, recv and recvfrom.

- *File access* functions: CreateFile, OpenFile, ReadFile and WriteFile.
- *Keyboard (Keys)* status functions: GetAsyncKeyState, GetKeyboardState, GetKeyNameText and keybd_event.

The communication functions are used because the bots needs to communicate with the botmaster in order to send or receive information. In addition, these function calls are used in flooding attack. The file access functions are needed because once a bot intercept the user keystrokes, it needs to store the intercepted data in a buffer or in a file for future access. The keyboard status functions are needed because many existing bots implement the keystrokes logging by executing these functions in `user mode' level in windows environment.

### 4.6 Experiments

The aim of these experiments is to evaluate the performance of the SRC algorithm and the DCA on detecting the bot running on the system. Various experiments are conducted to verify this aim. Each experiment is repeated ten times which is sufficient, as the results from the repeated experiments produce a small variation on standard deviation by using Chebyshev's Inequality. One dataset is selected randomly from each repeated experiments and is passed to both the SRC algorithm and the DCA. Five null hypotheses are used for the evaluation as shown in next section.

#### 4.6.1 Null Hypotheses

- Null Hypothesis One (H1): Data collected per dataset are normally distributed. The Shaprio-Wilk test is used for this assessment.
- Null Hypothesis Two (H2): The SRC algorithm is able to detect the existence of bot when correlating different attributes.
- Null Hypothesis Three (H3): The DCA algorithm using the MCAV/MAC values for the normal processes are not statistically different from those produced by the bot process. This is verified through the performance of a two-sided Mann-Whitney test.
- Null Hypothesis Four (H4): Variation of the signal weights in DCA algorithm as described in Table 2 produces no observable difference in the resultant MCAV/MAC values and the detection accuracy. Wilcoxon signed rank tests (two-sided) are used to verify this hypothesis.
- Null Hypothesis Five (H5): There is no difference between the SRC algorithm and DCA in terms of performance on detecting bot.

### 4.7 System Setup

In all DCA experiments, the parameters used are identical to those implemented in [14], with the exception of the weights. All experiments are performed in a small virtual IRC network on a VMware workstation. The VMware workstation runs under a Windows XP SP2 P4 2.4 GHz processor. The virtual IRC network consists of two machines, one IRC server and one infected host machine. Two machines are sufficient to perform these experiments as one host is required to be infected and the other to be an IRC server to issue commands to the bot in question. The statistical analyses are performed using R statistical computing package (v.2.6.0).

## 5. Results and Analysis

Upon the application of the Shapiro-Wilk test to each of the datasets, the resultant p-values imply that the distribution of the datasets is not normal. Therefore, the null hypothesis one (H1) is rejected. As a result of this, further tests with these data use non-parametric statistical tests such as the Mann-Whitney test, also using 95% confidence.

### 5.1 Spearman's Rank Correlation - SRC

Our assumption is that calling GetAsyncKeyState() or GetKeyboardState() functions by an unknown running program may represent abnormal behaviour in our system. This is because many of the current logging techniques in *user-mode* level in windows environment use these two function calls to perform keylogging activities. However, we consider that calling these functions generate only a "weak" alert because other legitimate programs may use the same API function calls. Therefore, the correlation of different types of bot behaviour is needed to enhance the detection confidence to form a "strong" alert.

In our experiments, we use the SRC algorithm to correlate two different datasets. The first dataset is PS and SS signals $(S_1, S_3)$ dataset while the second dataset is DS and SS signals $(S_2, S_3)$ dataset. In both datasets, we compare $S_1$ and $S_2$ with $S_3$ because the existing of $S_3$ suppresses the effect of other two signals.

We analyse the results of the experiments described in Section 4.2. Table 3 represents the SRC value between the two datasets, $(S_1, S_3)$ and $(S_2, S_3)$, in each experiment. In this table, we have two sets of results. In set *Set1*, we correlate all the captured data from our algorithm including the idle period. In this period, no activity is seen, therefore, we assign a zero value to this period. This is represented by (Z) columns. In set *Set2*, we remove all the idle periods which have zeros (NZ columns) and apply the SRC algorithm to the new data. The reason for having the two sets is that having the idle periods in our data increases the correlation value. This is because there are many places where no activity is noticed in both datasets, which may produce inaccurate correlation. Therefore, we wanted to investigate the effect of having no idle periods.

The *Keylogging Activity* column represents the situation where the process calls any function used to intercept the keystrokes. As a result, we classify our API detection confidence into three cases:

- Normal detection (Normal): Keylogging activity is not detected and either low or high correlation value is noticed.
- Weak detection (Weak): Keylogging activity is detected but a low correlation is noticed in both datasets.
- Medium detection (Medium): Keylogging activity is detected but a high correlation is noticed in one dataset.
- Strong detection (Strong): Keylogging activity is detected but a high correlation is noticed in both datasets.

As mentioned in section 3.1, a high correlation is considered if the SRC value exceeds the threshold (0.5). From Table 3,

if we consider Set2, we see a high correlation value between $(S_1, S_3)$ and $(S_2, S_3)$ in experiment E1. This is because the bot was inactive during all the time period. The only traffic generated by the bot is the PONG message to avoid disconnection from the IRC server. Therefore, the correlation value is expected to be high as well. We consider this situation as a "normal" case.

In experiment E2.1.a/b, the bot intercepts the user keystrokes and sends the data to the botmaster. As a result, a high correlation value is expected and "strong" detection is generated.

| Experiment | SRC($S_1$,$S_3$) | | SRC($S_2$,$S_3$) | | Keylog. Activities existence | API Detection Confid. |
|---|---|---|---|---|---|---|
| | Set1 (Z) | Set2 (NZ) | Set1 (Z) | Set2 (NZ) | | |
| E1 | 0.98 | 0.72 | 0.96 | 0.87 | No | Normal |
| E2.1.a | 0.61 | 0.85 | 0.74 | 0.69 | Yes | Strong |
| E2.1.b | 0.62 | 0.87 | 0.75 | 0.74 | Yes | Strong |
| E2.2.a | 0.64 | 0.51 | 0.60 | 0.59 | No | Normal |
| E2.2.b | 0.55 | 0.50 | 0.53 | 0.51 | No | Normal |
| E2.3.a | 0.11 | 0.17 | 0.50 | 0.52 | Yes | Medium |
| E2.3.b | 0.20 | 0.32 | 0.58 | 0.57 | Yes | Medium |
| E3 | 0.99 | 0.50 | 0.97 | 0.58 | No | Normal |

*Table 3.* The results of applying SRC on dataset signals $(S_1, S_3)$ and $(S_2, S_3)$.

In experiment E2.2.a/b, we notice a high correlation value on both datasets. This situation is expected because the attacker issues a SYN attack and a UDP attack. The bot responds by generating a large number of same communication function calls for a long period. No keylogging activity is detected during this period. As a result, a "normal" case is indicated. This situation represents the false negative case as it incorrectly classified as normal.

Experiment E2.3.a/b shows a combined keylogging and SYN/UDP attack activities. The correlation value of $(S_1, S_3)$ is low compared to experiment E2.2.a/b. This is because the bot is intercepting keystrokes and performing the SYN/UDP attack simultaneously. As a result, the two datasets were noisy which generate a "medium" detection case.

The last experiment E3 shows the result of applying SRC algorithm on the IceChat client. Even though we have a high correlation value before and after removing idle periods on both experiments, we did not detect the use of keylogging function calls. Notice that we do not have a "weak" scenario in this case.

In summary, we notice that some experiments produce low correlation values. There are many reasons for this. The first reason is that different events occur in different time-windows. As a result, SRC algorithm produces inaccurate results. The second reason is that some signals are varying differently influencing the correlation value. Meanwhile, we have many idle periods in our datasets, increasing the correlation value which affects our detection scheme. To improve this, we need to apply a more intelligent correlation scheme, as described in the next section. As a result, we can not reject or accept the Null Hypothesis Two (H2) as we need a strong correlation algorithm to perform a better indication of malicious behaviour.

### 5.2    DCA

The results from the DCA experiments are shown in Table 3 and Table 4. The mean MCAV and the mean MAC values for each process are presented, derived across the ten runs performed per scenario.

| Exper-iment | Pro-cess | Out-put Antgn | mean | | Mann-Whitney (p-value) | |
|---|---|---|---|---|---|---|
| | | | MCAV | MAC | MCAV | MAC |
| E1 | Bot | 35 | 0.0978 | 0.0578 | | |
| | IRC | 24 | 0.0625 | 0.0255 | **0.1602** | 0.0202 |
| E2.1.a | Bot | 1329.7 | 0.4736 | 0.4542 | | |
| | IRC | 59 | 0.2881 | 0.0122 | 0.0002 | 0.0002 |
| E2.1.b | Bot | 1296.2 | 0.5441 | 0.2098 | | |
| | IRC | 464.9 | 0.5284 | 0.1077 | 0.0089 | 0.0002 |
| | Cmd | 8.9 | 0.7889 | 0.0031 | 0.0002 | 0.0002 |
| | Note-pad | 239.4 | 0.6916 | 0.0726 | 0.0002 | 0.0002 |
| | Word-pad | 268.8 | 0.8286 | 0.0977 | 0.0002 | 0.0002 |
| E2.2.a | Bot | 19206 | 0.6047 | 0.6038 | | |
| | IRC | 18 | 0.3441 | 0.0003 | 0.0002 | 0.0000 |
| | cmd | 9.8 | 0.2889 | 0.0002 | 0.0003 | 0.0000 |
| E2.2.b | Bot | 5790.5 | 0.4360 | 0.4346 | | |
| | IRC | 19 | 0.2772 | 0.0009 | 0.0002 | 0.0000 |
| E2.3.a | Bot | 41456 | 0.8218 | 0.8214 | | |
| | IRC | 20.5 | 0.5480 | 0.0003 | 0.0002 | 0.0000 |
| E2.3.b | Bot | 22446 | 0.9598 | 0.9461 | | |
| | IRC | 59.1 | 0.7802 | 0.0021 | 0.0000 | 0.0002 |
| | Cmd | 9.7 | 0.6300 | 0.0003 | 0.0002 | 0.0002 |
| | Note-pad | 23.1 | 1.0000 | 0.0010 | 0.0001 | 0.0002 |
| | Word-pad | 233.6 | 0.8801 | 0.0090 | 0.0002 | 0.0002 |
| E3 | | 135.5 | 0.1136 | 0.1136 | N/A | N/A |

*Table 4.* The results of the MCAV/MAC values generated from DCA based on signal weights (WS$_3$). Values on bold font are not significant.

For all scenarios E1-E3, a comparison is performed using the results generated for the bot versus all other normal processes within a particular session as shown in Table 4. In this table, the computed p-values using an unpaired Mann-Whitney test are presented, with those results deemed not statistically significant marked in bold font. In experiment E1, no significant differences is noticed between the resultant MCAV values for the inactive bot and the normal IRC process, and so for this particular scenario the Null Hypothesis Three (H3) cannot be rejected for the reason of having small number of antigen produced by both processes to give an accurate description of the state of the monitored host. This is supported by the fact that the MAC values differ significantly for this experiment. This implies that the MAC is a useful addition to the analysis as it allowed for the incorporation of the antigen data, which can influence the interpretation of the results.

Significant differences are shown by the low p-values presented in Table 4 for experiments E2.1.a and E2.1.b for both the MAC and MCAV coefficient values, where the sample size is equal to ten. The differences are further pronounced in the generation of the MAC values, further supporting its future use with the DCA. We can conclude therefore, that the DCA can be used in the discrimination between normal and bot-directed processes and that the DCA is successful in detecting keylogging activities. This trend is also evident for scenarios E2.2.a/b and E2.3.a/b, where the bot process MCAV and MAC values are consistently higher than those of the normal processes, IRC and notepad inclusive. This information is also displayed in Figure 1 and Figure 2 respectively. This implies that in addition to the detection of the bot itself the DCA can detect the performance of outbound scanning activity. Therefore the Null Hypothesis Three (H3) can be rejected as in the majority of cases the DCA successfully discriminates between normal and bot processes, with the exception of E1 because of the extrusion approach that we are taking.

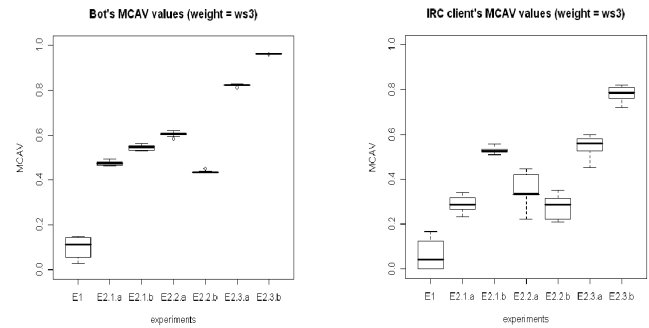

**Figure 1.** The MCAV values of bot and IRC client generated by DCA based on the weights (WS$_3$).

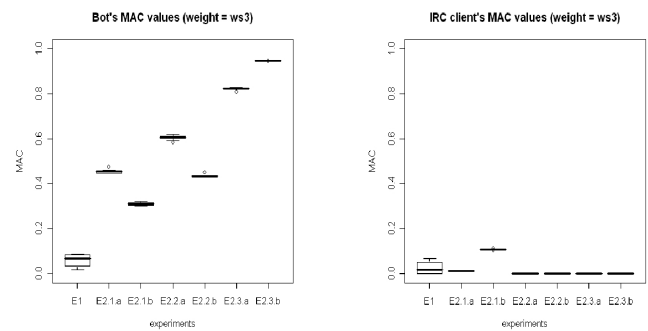

**Figure 2.** The MAC values of bot and IRC client generated by DCA based on the weights (WS$_3$).

Table 5 and Table 6 include the results of the sensitivity analysis on the weight values for the bot process. The aim of these experiments is to examine the effect of varying weight signals on to the DCA detection performance. Different values have been generated randomly to see the effect of increasing or decreasing S$_1$, S$_2$ and S$_3$ weight signals for O$_1$=csm, O$_2$=semi-mature and O$_3$=mature cell.

| Experiment | WS$_1$ | WS$_2$ | WS$_3$ | WS$_4$ | WS$_5$ |
|---|---|---|---|---|---|
| E1 | 0.05 | 0.08 | 0.10 | 0.11 | 0.14 |
| E2.1.a | 0.10 | 0.20 | 0.47 | 0.55 | 0.76 |
| E2.1.b | 0.38 | 0.41 | 0.54 | 0.59 | 0.80 |
| E2.2.a | 0.55 | 0.31 | 0.60 | 0.93 | 0.93 |
| E2.2.b | 0.30 | 0.18 | 0.43 | 0.59 | 0.94 |
| E2.3.a | 0.88 | 0.63 | 0.82 | 0.91 | 0.99 |
| E2.3.b | 0.95 | 0.94 | 0.95 | 0.96 | 0.99 |

*Table 5.* Weight sensitivity analysis for the bot's MCAV values.

For example, in case of O$_1$, we have increased and decreased the weight value of $S_1$, $S_2$ and $S_3$ to the point that reaches the steady state where further increase and decrease to these values will not have a large impact on the MCAV/MAC values.

| Experiment | $WS_1$ | $WS_2$ | $WS_3$ | $WS_4$ | $WS_5$ |
|---|---|---|---|---|---|
| E1 | 0.03 | 0.05 | 0.06 | 0.07 | 0.08 |
| E2.1.a | 0.09 | 0.19 | 0.45 | 0.52 | 0.73 |
| E2.1.b | 0.22 | 0.23 | 0.31 | 0.33 | 0.46 |
| E2.2.a | 0.55 | 0.31 | 0.60 | 0.92 | 0.93 |
| E2.2.b | 0.29 | 0.18 | 0.43 | 0.58 | 0.94 |
| E2.3.a | 0.88 | 0.63 | 0.82 | 0.91 | 0.99 |
| E2.3.b | 0.94 | 0.93 | 0.95 | 0.95 | 0.97 |

*Table 6.* Weight sensitivity analysis for the bot's MCAV values.

The values presented in Tables 5 and 6 are mean values taken across the ten runs per session (E1-E2). An arbitrary threshold is applied at 0.5; values above this threshold deem the process anomalous, and below as normal. From these data, it is shown that changing the weights used in the signal processing equation has significant effect on the performance of the system. For example, in the case of session E2.1.a, weight set $WS_1$ produces a MAC value of 0.09 for the bot yet produces a value of 0.73 for $WS_5$. This increase is likely to reduce the rate of false negatives. To further explore these effects, the resultant data are plotted as boxplots as the data are not normally distributed. To assess the performance of the DCA as an anomaly detector the results for the anomalous bot and the normal IRC client are shown for the purpose of comparison. For these boxplots, the central line represents the median value, with the drawn boxes representing the interquartile ranges.

In Figure 3 the median MCAV values are presented, derived per session across the ten runs performed for each WS (n=50).

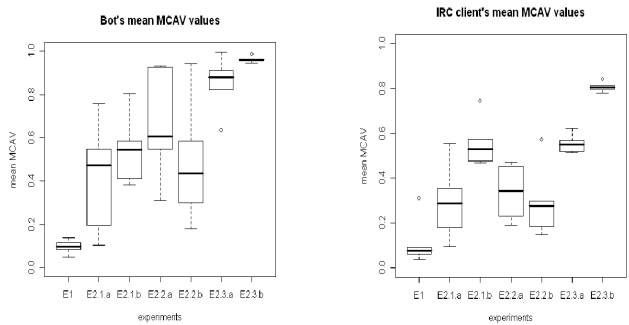

**Figure 3.** The mean MCAV values for the bot and IRC client generated by DCA using different signal weight values ($WS_1$–$WS_3$).

For the bot process, the MCAV is low for session E1, in-line with previous results. For E1, variation in the weights does not influence the detection results, as this process has low activity and therefore does not generate any great variation in the signals. Therefore, without input variation, the output does not vary in response to changing the manner in which the input is processed. This is also evident in Figure 4 when using the bot's MAC values.

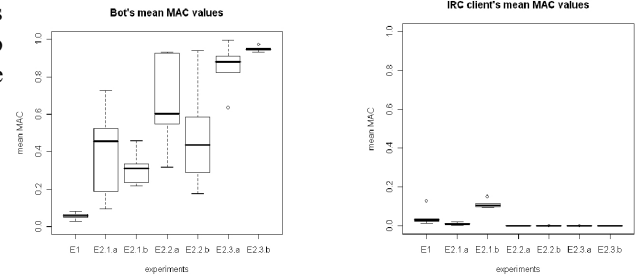

**Figure 4.** The mean MAC values for the bot and IRC client generated by DCA using different signal weight values ($WS_1$–$WS_3$).

For all other sessions, much greater variation is observed upon weight modification, as shown by the large interquartile ranges produced for both MCAV and MAC values of the bot processes. While the similar trends are shown across the sessions in the MCAV of the IRC client, differences are evident for the MAC value. In Figure 4 (Bot's mean MAC values) it is evident that all sessions have low MACs for this process across all weight sets. Therefore as the weights are modified, there is a greater influence on the anomalous processes than on the normal processes. Should the arbitrary threshold applied to the MAC values be set at 0.2 as opposed to 0.5, then the performance of the DCA on botnet detection is good, producing low rates of false positives and high rates of true positives.

Finally, to verify these findings statistically, each set of results per session per weight are compared exhaustively using the non-parametric Wilcoxon signed rank test. For each test performed the resultant p-value is less than 0.001. This allows us to conclude that modification of the weights has a significant effect on the output of the DCA when applied to this detection problem, and leads to the rejection of Null Hypothesis Four (H4).

### 5.3 SRC Algorithm and the DCA Performance

From the results obtained, even though that both algorithms were able to detect the malicious behaviours by correlating different attributes, we notice that the DCA has a better performance over the SRC algorithm when detecting the bot by reducing the number of false alarms and classifying processes into normal and malicious. Therefore, the Null Hypothesis Five (H5) can be rejected.

## 6. Conclusion

In this work, we try to evaluate the performance of two correlation algorithms on bots detection by correlating different activities which inhibits malicious behaviour. After collecting our datasets, we pass the captured data to a Spearman's rank correlation (SRC) algorithm. Although SRC algorithm is a simple method to examine the correlation level, the results were promising. However, some experiments show a low correlation values. This is because different activities occur in different time-windows. As a result, high false negative values could be generated.

We applied the same datasets to the DCA to evaluate its detection accuracy and performance in comparison to SRC

by exploring different null hypotheses. It is shown that the DCA is capable of discriminating between bot and normal processes on a host machine. Additionally, the incorporation of the MAC value has a significantly positive effect on the results, significantly reducing false positives. Finally, the modification of the weights used in the signal processing component has a significant effect on the results of the algorithm. In addition, we noticed that appropriate weights for this application include high values for the safe signal

## Acknowledgment

The authors would like to thank Khalifa University of Science, Technology And Research (KUSTAR) - UAE, for providing financial support for this work.

(SS) weight which appears to be useful in the reduction of potential false positives without generating false negative errors. We can conclude that the performance of correlating different activities using DCA is better than SRC algorithm. We are now aiming to apply the DCA to the detection of "peer-to-peer" bots, which pose an interesting problem as the use of peer-to-peer networks increases.

## Author Biographies

**Yousof Al-Hammadi** is a PhD student in school of computer science and information Technology at the University of Nottingham. He received his Bachelor degree in Computer Engineering from Etisalat College of Engineering, UAE in 2000 and a MSc in Telecommunications Engineering from the University of Melbourne, Australia in 2003. His research interests focus on detecting Internet worms, IRC botnet/bots using anomaly detection techniques where he applies the Dedritic Cell Algorithm (DCA) in order to detect IRC bots running on a system. Currently, He focuses on applying the DCA on Peer-to-Peer (P2P) bots.

**Uwe Aickelin** received a Management Science degree from the University of Mannheim, Germany, in 1996 and a European Master and PhD in Management Science from the University of Wales, Swansea, UK, in 1996 and 1999, respectively. He worked in the Mathematics Department as a lecturer in Operational Research at the University of the West of England in Bristol. In 2002, he accepted a lectureship in Computer Science at the University of Bradford. Since 2003 he works for the University of Nottingham in the School of Computer Science where he is now a Professor of Computer Science and leader of the IMA group.

Prof. Aickelin currently holds an EPSRC Advanced Fellowship focusing on AIS, anomaly detection and mathematical modelling. He has been awarded an EPSRC research funding as Principal Investigator on topics including AIS, Danger Theory, Computer Security, Robotics and Agent Based Simulation. Prof. Aickelin is an Associate Editor of the IEEE Transactions on Evolutionary Computation, the Assistant Editor of the Journal of the Operational Research Society and an Editorial Board member of Evolutionary Intelligence.

**Julie Greensmith** is a Lecturer at the University of Nottingham. She gained a BSc in Pharmacology from the University of Leeds, UK in 2002 and a MSc in Multidisciplinary Informatics in 2003, also from the University of Leeds and completed a PhD in Computer Science at the University of Nottingham in 2007.Her research focuses on the development of novel AIS algorithms applied to computer security and bio-sensing for the entertainment industry.